\documentclass{article}

\usepackage[final]{gpsmdms_2022}

\usepackage[utf8]{inputenc} 
\usepackage{hyperref}       
\usepackage{url}            
\usepackage{booktabs}       
\usepackage{amsfonts}       
\usepackage{nicefrac}       
\usepackage{microtype}      
\usepackage[dvipsnames]{xcolor}         
\usepackage{mathtools}

\hypersetup{
    colorlinks=true,
    linkcolor=MidnightBlue,
    urlcolor=MidnightBlue,
    citecolor=MidnightBlue,
    }

\usepackage{amsmath, amssymb}
\usepackage{graphicx}
\usepackage{floatrow}
\usepackage{subfig}
\usepackage{caption}
\usepackage[
    backend=biber,
    natbib=true,
    sorting=none,
    doi=false,
    maxcitenames=2,
    ]{biblatex}
\usepackage{chngcntr}

\newcommand{\stwocvi}{S$^{2}$CVI}
\newcommand{\term}[1]{\textit{#1}}

\title{Spatiotemporal modeling of European paleoclimate using doubly sparse Gaussian processes}

\author{%
    Seth D.~Axen$^{1}$\thanks{E-mail: \href{mailto:seth@sethaxen.com}{seth@sethaxen.com}} \quad Alexandra Gessner$^{1}$\quad Christian Sommer$^2$\\ \quad \textbf{Nils Weitzel}$^3$ \quad \textbf{Álvaro Tejero-Cantero}$^1$ \\

    $^1$Cluster of Excellence Machine Learning\\
    $^2$The Role of Culture in Early Expansions of Humans,\\ Heidelberg Academy of Sciences and Humanities\\
    $^3$Department of Geosciences\\
     University of Tübingen, Germany
}

\begin{document}

\maketitle

\begin{abstract}
    Paleoclimatology---the study of past climate---is relevant beyond climate science itself, such as in archaeology and anthropology for understanding past human dispersal.
    Information about the Earth's paleoclimate comes from simulations of physical and biogeochemical processes and from proxy records found in naturally occurring archives. 
    Climate-field reconstructions (CFRs) combine these data into a statistical spatial or spatiotemporal model.
    To date, there exists no consensus spatiotemporal paleoclimate model that is continuous in space and time, produces predictions with uncertainty, and can include data from various sources.
    A Gaussian process (GP) model would have these desired properties; however, GPs scale unfavorably with data of the magnitude typical for building CFRs.
    We propose to build on recent advances in sparse spatiotemporal GPs that reduce the computational burden by combining variational methods based on inducing variables with the state-space formulation of GPs.
    We successfully employ such a doubly sparse GP to construct a probabilistic model of European paleoclimate from the Last Glacial Maximum (LGM) to the mid-Holocene (MH) that synthesizes paleoclimate simulations and fossilized pollen proxy data.
\end{abstract}

\section{Introduction}
Paleoclimate reconstructions are important for understanding climate processes and variability but are also valuable to other scientific domains such as archaeology or paleoecology.
For example, past climate variability has been a driver for human migration, adaption, and cultural innovation \cite{timmermann2022climate,timmermann2016late, demenocal2011climate}.
For reconstructing the climate of the past, there are two main sources of information.
The first source is the known dynamics of the climate system encoded as a climate model.
Paleoclimate states are generated from these models through numerical simulations under reconstructed or theorized boundary conditions called forcings.
The simulation outputs are typically represented as a spatial grid of values of climate variables like mean annual temperature or total annual precipitation.
Some simulations estimate climate variables at one or a few relevant time slices, while others do so at regular intervals (Fig.~\ref{fig:figa1}).
The second source of information is proxy data from geolocated natural archives that contain information about the time of their deposition, where the proxy data informs the corresponding climate state at that time (Fig.~\ref{fig:figa1}).
Such archives include ice cores, mineral deposits in caves, and lake sediment cores.
Fossilized pollen grains, found within different types of archives, are the most abundant terrestrial proxy due in part to their comparatively high durability.
These pollen grains are classified into taxonomic groups and compared to modern species distributions.
Relating the modern taxa to modern climate allows one to generate paleoclimate reconstructions with existing techniques \citep{bartlein2011pollen, ohlwein2012review, chevalier2020pollen}.
Because pollen are found at archaeological sites worldwide, they are particularly compelling for studying the environmental context of human evolution. 

\term{Climate field reconstructions} (CFRs) infer spatial or spatiotemporal fields of climate variables with statistical models.
Most Bayesian CFRs consider only the more data-rich recent past, i.e. the last two thousand years (2 ka), and most of these use information only from proxies and not from simulations \cite{tingley2012piecing,hakim2016last}.
Purely spatial probabilistic reconstructions of paleoclimate from simulations and pollen have been built by \citet{weitzel2019combining, weitzel2022state}. 
\citet{osman2021globally} used ensemble Kalman filtering to model surface temperature based on multiple simulation time slices from one climate model and sea surface temperature proxy data from the LGM to the present; this study did not explicitly model temporal autocorrelations.
To our knowledge, the only fully Bayesian spatiotemporal model for CFR of paleoclimate has been presented by \citet{weitzel2020diss} but applied only to synthetic data.
Their parametric hierarchical model has $\mathcal{O}(10^5)$ parameters, which makes inference challenging.

\looseness=-1
Currently there is no consensus CFR available that is global, spatiotemporal, probabilistic, uses the majority of available proxy data and simulations, and can be queried at an arbitrary location and time.
Our goal is ultimately to construct such a model and make it available to the research community.
As a first step, we present here a spatiotemporal Gaussian process model of mean annual temperature across Europe from the \term{Last Glacial Maximum} (LGM; $\sim$ 21 ka) to the \term{mid-Holocene} (MH; $\sim$ 6 ka) based on simulations from various paleoclimate models and reconstructions from fossilized pollen (cf. Section~\ref{sec:model}).
Inference on such a large dataset ($N>6\cdot 10^5$, cf. Section~\ref{sec:data}) is made tractable using a combination of inducing variables and a Markovian structure in the temporal domain \citep{adam2020doubly, wilkinson2021sparse}, using \term{doubly sparse} GPs (cf. Section~\ref{sec:inference}).
With only 100 spatial and six temporal inducing points, our model interpolates well in space and time between data points and comes with calibrated uncertainty estimates.

\section{Dataset}
\label{sec:data}

In this work we focus on mean annual temperature as the climate variable of interest using fossilized pollen grains as proxy data.
We constrain the spatial domain to the European continent during the period from LGM to MH.
All data is from publicly available databases.

\paragraph{Pollen data} We use 41,986 site-specific reconstructions of mean annual temperature from fossilized pollen collected at 826 sites and 1,607 time slices from the \texttt{LegacyClimate$\!$ 1.0} database \cite{herzschuh2022_legacyclimate}.

\paragraph{Simulation data} We use a total of ten models with different output coverage in space and time.
Three models have been simulated only at MH \cite{voldoire2013_cnrmcm5,dufresne2013_ipslcm5alr,wu2019_bcccsm1}, while five are available at both LGM and MH but no other time slices \cite{gent2011_ccsm4,hadgem2,yukimoto2012_mricgcm3,sueyoshi2013_mirocesm,jungclaus2014_mpiesmp}.
All of the above have a spatial grid spacing of 10 arcminutes ($<15$~km) and were downloaded from WorldClim 1.4 \cite{hijmans2005_worldclim}.
Two models have been simulated at time slices spaced by 1,000 years ranging from LGM to MH and have a grid spacing of 30 arcminutes ($<46$~km) \cite{beyer2020,krapp2021}.

Simulations and pollen-derived reconstructions provide a total of $N=661,028$ data points.
We preprocessed the data by using radial basis functions to spatially interpolate within each lower-resolution MH simulation; these interpolations were then averaged to form a function $m(x)$.
This function $m(x)$ was then subtracted from all data; this smoothed out the steep gradients in climate variables around mountain ranges and the coastline.
Thus the data we use during modeling is the deviation from this empirical model.

\section{Description of probabilistic model}
\label{sec:model}

\paragraph{Prior} Let $C(x, t)$ be a climate variable of interest at coordinates $x = (\mathrm{longitude}, \mathrm{latitude})$ and time $t$.
We adopt the following zero-mean, factorized GP prior for $\overline{C}(x, t) \coloneqq C(x, t) - m(x)$:
\[
    \overline{C}(x, t) \sim \mathcal{GP}(0, k_x(x, x') k_t(t, t')),
\]
where $k_x$ is a Mat\'ern-\nicefrac{3}{2} spatial kernel with different latitudinal and longitudinal length scales, and $k_t$ is a Mat\'ern-\nicefrac{1}{2} temporal kernel corresponding to an Ornstein-Uhlenbeck process, whose state-space representation has dimension  $d=1$.

\paragraph{Likelihood} Let $Y_{s,p}$ be either a site-specific reconstruction or a simulated value at site $s$ with coordinates $x_s$ and time $t_p$.
Furthermore, let $\overline{Y}_{s,p} = Y_{s,p} - m(x_s)$.
We approximate for now that all data are independently and identically (i.i.d.) normally distributed
\[
    \overline{Y}_{s,p}\, |\, \overline{C}(x_s, t_p) \sim \mathcal{N}\big(\overline{C}(x_s, t_p), \sigma^2\big).
\]
The same $\sigma$ is used for all simulations as well as the pollen-based reconstructions.

This model yields an exact GP posterior for $\overline{C}$, which allows the resulting posterior distribution of $C$ to be queried at any spatiotemporal point $(x, t)$.
However, the large number of data points $N$ makes the $\mathcal{O}(N^3)$ inversion of the kernel Gram matrix intractable.
Therefore, we turn to a sparse GP approximation for inference.

\section{Approximate inference}
\label{sec:inference}

\paragraph{Doubly sparse spatiotemporal Gaussian processes}
To keep inference computationally feasible, we use a sparse variational GP approximation that is compatible with the time-series structure of our model.
Such algorithms are known as doubly sparse GP approximations because they combine two approaches to obtain sparsity: 1) a set of $M \ll N$ inducing variables and 2) the state-space representation of Markovian GPs, which allows for linear-time inference using Kalman updates \cite{adam2020doubly}.
With a variational distribution on the inducing states that retains the chain structure, the marginal posterior predictions can be computed efficiently using filtering methods.
For spatiotemporal models, these methods require a kernel that is separable into the product of a spatial kernel and a Markovian temporal kernel.
Among multiple recent algorithmic approaches to sparse Markovian GPs \cite{wilkinson2021sparse} we choose \stwocvi~(doubly sparse conjugate-computation variational inference) as implemented in \texttt{markovflow} \cite{markovflow}.
\stwocvi~defines a variational posterior for $\overline{C}$ that is optimized using natural gradient updates and has been reported to be more efficient and numerically stable than alternative approaches \cite{wilkinson2021sparse}.
The time complexity of evaluating the variational objective and its gradient using \stwocvi~ is $\mathcal{O}\bigl((M_t + N_b) (M_s d)^3\bigr)$, for $M_s$ spatial inducing points, $M_t$ temporal inducing points, dimension $d$ of state-space representation of $k_t$, and batch size $N_b$ \cite{wilkinson2021sparse}.
This method has not been used previously for problems of the scale of ours, and no public code exists of applications such as geospatiotemporal modeling where there is more than one spatial dimension.

\paragraph{Experimental details}
We trained \stwocvi~with $M_s=100$ spatial and $M_t=6$ temporal inducing points over 30 epochs with a minibatch size of $N_b=1,000$.
At each iteration, the parameters of the variational posterior are first updated using natural gradients, and then all hyperparameters are updated using Adam \cite{kingma2014adam}.
The coordinates of the spatial inducing points, which are constrained to stay within the spatial domain of the data, are treated as hyperparameters.
All predictions from the variational posterior for $\overline{C}$ were shifted by $m(x)$ to obtain the approximate posterior for $C$, and these are used in all subsequent analyses.

\section{Results}

\begin{figure}
    \centering
    \sidesubfloat[]{\label{fig:fig1a}\includegraphics[width=0.42\textwidth]{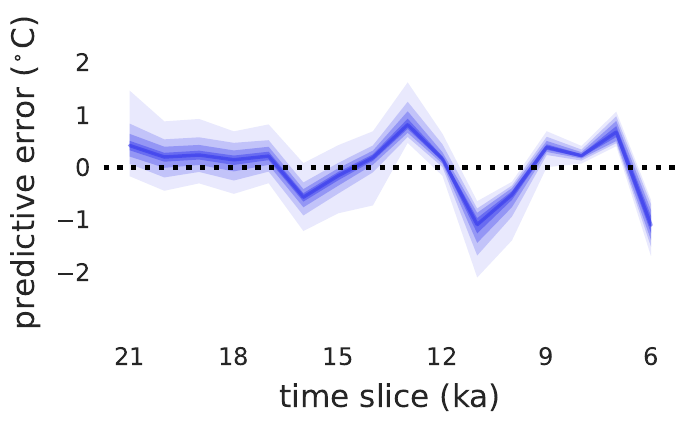}}
    \hfill
    \sidesubfloat[]{\label{fig:fig1b}\includegraphics[width=0.42\textwidth]{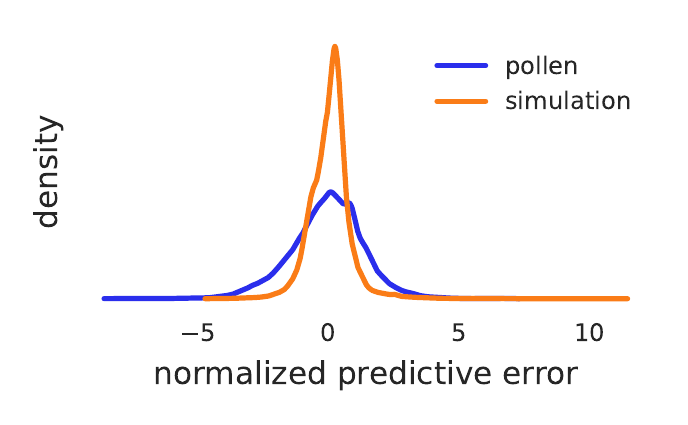}}
    \caption{
        Predictive error of model $C(x,t)$.
        \textit{(a)} Results of validation when fitting the model to a single multi-time step simulation \protect{\cite{krapp2021}}.
        For each model fit, a single time slice was held out from the simulation during training and compared with the corresponding posterior predictive (PP) means at the same time slice and coordinates.
        Error is computed as (PP mean $-$ data).
        Shown are the central intervals of the error distributions containing 80\%, 60\%, 40\%, and 20\% of values, as well as the median.
        \textit{(b)} Kernel density estimate (KDE) of normalized posterior predictive (PP) error distribution.
        Error is computed as (PP mean $-$ data) / (PP std).
        The KDE is computed using ArviZ \protect{\cite{arviz}}.
    }
    \label{fig:fig1}
\end{figure}

\paragraph{Validation} We validated the general approach by performing ``leave-one-time-slice-out'' inference with a single multi-time slice simulation \cite{krapp2021} as data (Fig.~\ref{fig:fig1a}).
The average posterior predictive error comparing each posterior predictive mean to the corresponding measurement in the held-out time slice across all time slices is 0.05 ${}^\circ$C with a 95\% central posterior predictive interval of (-1.86 ${}^\circ\text{C}$, 2.24 ${}^\circ\text{C}$).
The mean absolute posterior predictive error is 0.69 ${}^\circ$C.

\begin{figure}
  \centering
  \includegraphics[width=0.9\linewidth]{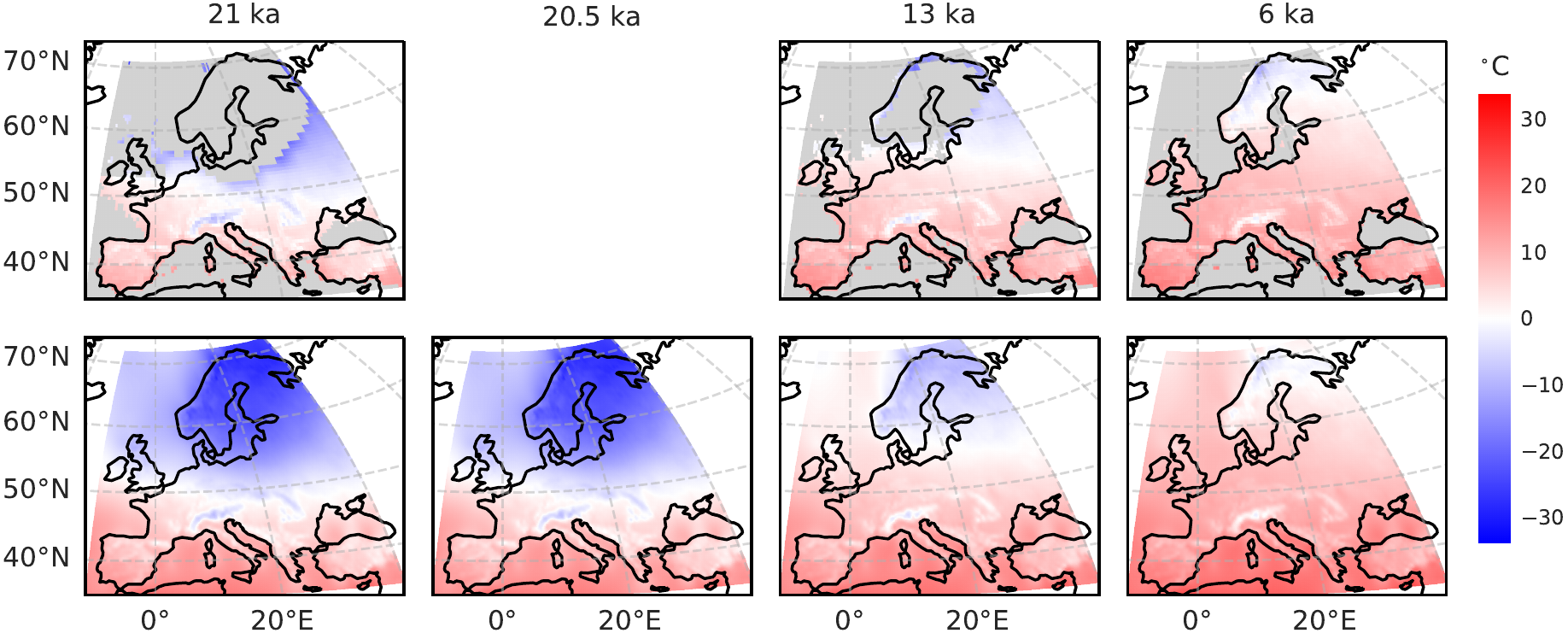}
  \caption{
      Comparison of mean annual temperature from one of the regularly temporally spaced simulations \protect{\cite{krapp2021}} ($Y$; \textit{top}) with the mean of the variational posterior approximation ($C(x,t)$; \textit{bottom}) at several time slices.
      At $20.5\,\mathrm{ka}$, there is no simulated data available, so only the model's prediction is shown.
      The grey area in the simulation indicates grid points that are masked due to presence of ocean or an ice sheet.
      Because no observations are available in these regions, the model's predictions there are not expected to be reliable.
  }
  \label{fig:fig2}
\end{figure}

\paragraph{Inference} Training on the entire dataset took 36 hours on an NVIDIA V100 GPU.
98.9\% of all data fell within three standard deviations of the mean of the posterior predictive distribution (99.1\% for simulations and 95.9\% for pollen; Fig.~\ref{fig:fig1b}).
The mean of the variational posterior both spatially and temporally interpolates between the simulations (Fig.~\ref{fig:fig2}).

\section{Outlook}

\looseness=-1
We are working on several immediate improvements to the model.
Instead of using a weighted interpolation of simulations as an empirical model to be subtracted from data, we will use a spatial interpolation of measurements from the early 1900s. 
To allow capturing remaining large-scale spatial and temporal trends, we will further introduce a non-zero parametric prior mean function to the prior on $\overline{C}$.
We will also incorporate prior beliefs about the values of the fitted parameters by setting weakly informative priors.
As a step toward extending the model to accommodate different data sources, we will use separate likelihood functions for proxy data and simulations.
In the medium term, we would like to drop the i.i.d. assumption for simulations by including a sparse covariance matrix to account for the spatial covariance of the simulated data.
Moreover, climate variables often co-vary, so we plan to jointly model mean annual temperature and total annual precipitation using a multi-output GP.
Dating of proxies has its own uncertainty, which is especially relevant for pollen and is currently unaccounted for in our model; another extension would be to include this uncertainty in the variational objective \cite{damianou2016inputs, van2020framework}.
In the future we will expand the geographical extent of the model to provide predictions of climate with global land coverage between LGM and MH.

\section{Conclusion}

Paleoclimate modeling lacks consensus spatiotemporal models that combine simulations and proxies and adequately report uncertainty estimates.
Progress in this area is hindered by the computational expense of training such models.
We demonstrate that state-of-the-art sparse GP algorithms enable spatiotemporal modeling in realistic applications in the climate sciences where large datasets of different modalities are common.
While the result already possesses the desired properties for a consensus model, we describe model improvements that will enhance its usefulness for geosciences communities.
Our paleoclimate synthesis will ultimately be made available as a free, queryable web app to facilitate discovery in applied fields such as archaeology and paleoecology.

\begin{ack}
    Funded by the Deutsche Forschungsgemeinschaft (DFG, German Research Foundation) under Germany’s Excellence Strategy – EXC number 2064/1 – project number 390727645.
    NW acknowledges funding by the DFG, project number 395588486.
    CS acknowledges funding by the ROCEEH project (The Role of Culture in Early Expansions of Humans) of the Heidelberg Academy of Sciences.
    We thank Vincent Adam and Elizaveta Seminova for useful discussions.
\end{ack}

\section*{References}

\printbibliography[heading=none]


\clearpage

\appendix

\counterwithin{figure}{section}

\section{Results of hyperparameter optimization}

\begin{figure}[h]
  \centering
  \includegraphics[width=0.9\linewidth]{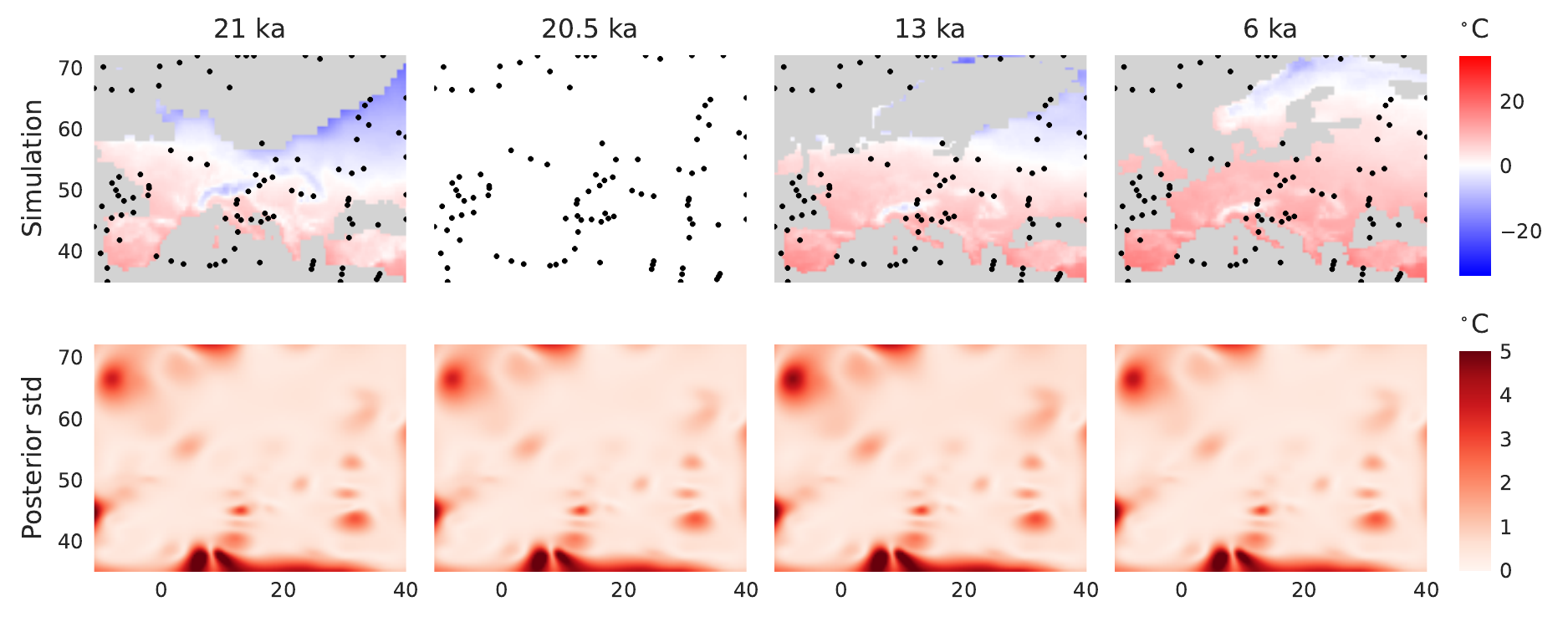}
  \caption{
      Comparison of mean annual temperature from one of the regularly temporally spaced simulations \protect{\cite{krapp2021}} (\textit{top}) with the standard deviation of the variational posterior approximation (\textit{bottom}).
      The black dots in the top row are the optimized locations of the spatial inducing points, which are shared across all time slices.
  }
  \label{fig:figa2}
\end{figure}

The spatial kernel of the GP prior $k_x$ has two length scales in units of degrees: a longitudinal one $\ell_\mathrm{lon}$ and a latitudinal one $\ell_\mathrm{lat}$.
After optimization, these took the values $\ell_\mathrm{lon}=19.6^\circ$ and $\ell_\mathrm{lat}=13.2^\circ$.
The temporal kernel $k_t$ has a single length scale $\ell_t$, which was optimized to $\ell_t = 9,900$ years.
The global standard deviation of the product kernel $k$ was optimized to 2.9 $^\circ$C.
The standard deviation $\sigma$ of the likelihood was optimized to a value of 1.6$^\circ$C.
The optimized spatial inducing point locations are shown in Fig.~\ref{fig:figa2}.

\section{Overview of data and consensus model}

\begin{figure}[hbt]
  \centering
  \includegraphics[width=0.8\linewidth]{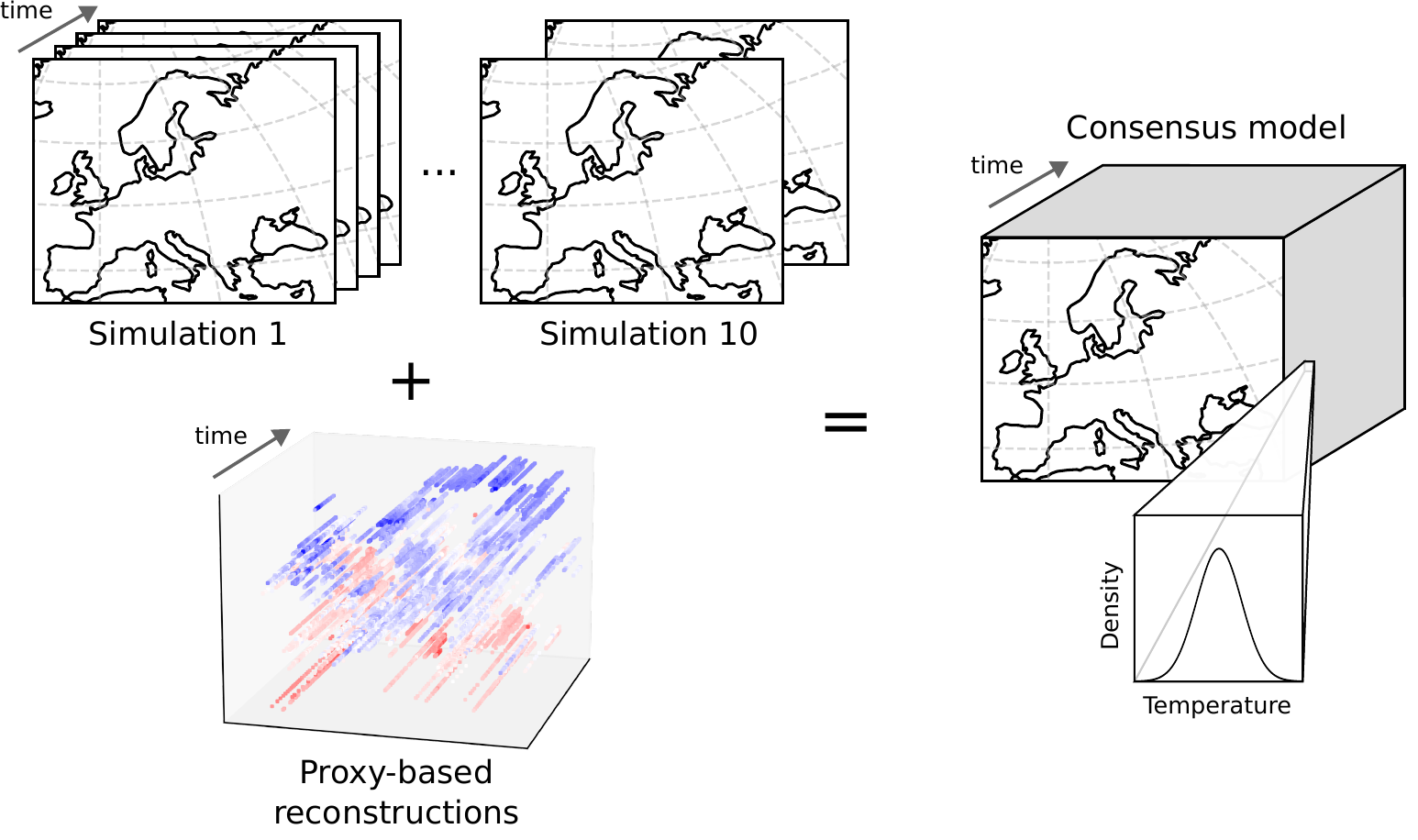}
  \caption{
        Overview of data and consensus model.
        All colors represent temperature.
        For proxy-based reconstructions (\emph{bottom}), each marker represents a reconstruction at a single spatiotemporal point.
        Multiple reconstructions at different points in time within a single archive appear stacked.
  }
  \label{fig:figa1}
\end{figure}

Multiple spatiotemporally gridded simulations are combined with reconstructions from fossilized pollen proxies to construct a consensus model (Fig.~\ref{fig:figa1}).
While the data are discrete in space and time, the consensus model is continuous in both space and time, and at any spatiotemporal point, the marginal posterior distribution of temperature can be queried. 

\end{document}